# CliffRank: A Dual-Branch Framework for Activity-Cliff Ranking Prediction

Kewei Li[1,2], Rongying Zhang[3], Peiyu Yang[1,2], Zhongjian Wang[4], Qiuchen Zhao[5,6], Lan Huang[1,2], Fengfeng Zhou[1,2,#].

1 College of Computer Science and Technology, Jilin University, Changchun, 130012, China.

2 Key Laboratory of Symbolic Computation and Knowledge Engineering of Ministry of Education, Jilin University, Changchun, 130012, China.

3 Greenwich High School, Greenwich, CT, 06830, United States of America.

4 BCPM Data Limited, Chengdu 610041, China.

5 Cancer Research UK Cambridge Centre and Department of Oncology, University of Cambridge, Cambridge, CB2 0XZ, UK

6 Department of Pathology, University of Cambridge, Cambridge, CB2 1QP, UK

# Correspondence may be addressed to Dr. Fengfeng Zhou (FengfengZhou@gmail.com or ffzhou@jlu.edu.cn).

## Abstract

Activity-cliff ranking remains difficult because local structural changes can cause large activity differences, while high-quality data that resolve the underlying mechanisms remain limited. To use available activity labels more effectively, we combine absolute-activity regression with ranking-consistency learning. CliffRank trains two parallel predictors with mean squared error, a thresholded listwise loss, and Pairwise Preference Consistency (PPC), which aligns relative ordering in the preference-probability space. On three antimicrobial peptide datasets, CliffRank with ESM2-t12 achieved the highest mean Spearman correlation of 0.5393 and mean Recall@50 of 21.4, although the leading method varied across individual datasets. On three small-molecule datasets, CliffRank with PNA, where PPC was activated after 120 epochs, achieved the highest mean Spearman correlation of 0.6890, while its mean Recall@50 of 30.4 matched that of ACANet-PNA. The PPC results also define its practical limits. Asymmetric initialization improved the MolCLR-GIN averages but did not improve every target. For PNA without pretrained weights, delayed PPC improved selected metrics, but no schedule was best for both mean Spearman correlation and mean Recall@50. Future work should evaluate more targets and antimicrobial peptide systems, develop adaptive PPC schedules, and incorporate protein or membrane context when available.

## Introduction

On unseen MMP pairs, QSAR models still struggle to identify activity cliffs (ACs) and predict activity differences[1, 2]. AC analysis is also relevant to identifying risks associated with local structural modifications during lead optimization[3]. AC formation depends on the target and structural context[3, 4]. Potential explanations include changes in binding modes, protein-ligand interactions, solvation or desolvation, entropy, and experimental error[1, 4-6]. Some membrane-active AMPs can self-assemble or oligomerize on membranes and increase membrane permeability through pore formation or other forms of membrane disruption[7]. These processes depend on dynamic peptide-peptide and peptide-membrane interactions. Static structural information alone cannot capture the full mechanism[7, 8]. MD studies have identified several modes of AMP-membrane interaction[8-10]. Experimental and theoretical studies also support related mechanistic models[7, 11]. However, mechanistic interpretation remains limited by accessible sampling timescales and simplified membrane models. Real membranes have complex compositions and lateral heterogeneity, which further limit the generalizability of simplified models[12]. For small molecules, local structural changes can alter binding through stereochemistry, conformational selection, interactions within or outside the pocket, and solvation[6, 13-16]. Binding-site water molecules and water networks can reorganize after ligand binding or local substitution. Such reorganization can alter binding thermodynamics or kinetics[17-20]. From a modeling perspective, small-molecule AC prediction is affected by training-set coverage, experimental error, and limitations of structural representations[2, 5]. Studies of AMP-membrane mechanisms are constrained by conformational sampling, timescale, and membrane-model complexity[10, 12].

Direct improvement of the data remains difficult because sufficiently large, high-quality datasets are hard to obtain. Public AMP-membrane MD studies usually focus on specific peptides and membrane systems. Complex, biologically realistic membrane models still pose substantial computational and sampling challenges[8, 10]. Simulating actual cellular membrane compositions is particularly difficult. We found no large, standardized benchmark of AMP-membrane trajectories that can be used directly for supervised AMP activity modeling. Protein-ligand structural data are also noisy. Small-molecule ligands in the PDB vary considerably in their fit to experimental density and in chemical geometry quality[21]. PDBbind may also contain protein or ligand structural artifacts and affinity measurements collected from heterogeneous sources[22]. Furthermore, high protein or ligand similarity between the training and test sets in the original splits can cause data leakage[23]. For computational modeling, data quality is generally more important than sheer data volume. We should therefore extract as much information as possible from the available data. Current representation-learning methods have not yet exhausted the information contained in these datasets.

AMPCliff was the first study to establish a quantitative definition and a systematic benchmark for ACs in AMPs composed of canonical amino acids. It also introduced the AC split to test whether models can generalize from the training data to unseen AMP cliff pairs[24]. The study reported both Spearman correlation and top-50 Recall. These metrics assess overall consistency in MIC ranking and the retrieval of highly ranked AMPs, respectively[24]. We adopt the same data splits and metrics to maintain comparability with the existing benchmark.

Studies in computer vision suggest that late-layer representations vary with the training paradigm and terminal training behavior[25, 26]. Because Spearman correlation is our primary metric, we further examine whether learning-to-rank objectives improve activity ranking. Prior work has shown that this approach can improve small-molecule prediction[27]. However, it has not been validated for AMP regression.

Learning-to-rank methods are commonly divided into pointwise, pairwise, and listwise approaches[28]. Pointwise methods treat each sample independently as a regression or classification instance. They can use absolute activity labels but do not directly encode the order between samples[29]. Pairwise methods use sample pairs as training instances and optimize their relative order. RankNet is a representative example[29]. Listwise methods treat an entire list as one training instance and learn the overall ranking through list probabilities or list-level losses. Examples include ListNet[30] and ListMLE[31]. Previous work has shown that pairwise loss can improve ranking regression[27]. However, in our AMP experiments, pairwise training caused large oscillations in the training loss. We hypothesize that pairwise supervision repeatedly pulls the same sample in different directions across different pairs. Because this study emphasizes the overall Spearman ranking, a listwise loss is better aligned with the evaluation objective. Our experiments also showed that the listwise loss stabilized the training loss.

Previous studies have also shown that neural networks with different initializations learn some shared features, but their representations are not identical[32]. Even with the same architecture, networks initialized with different random seeds may learn common features, features that are not consistently reproduced, or distinct representation bases[32]. JoCoR[33], a method for weakly supervised learning, jointly trains two networks for classification with noisy labels. It adds an inter-network prediction-consistency regularizer to the supervised classification losses of both networks. The resulting joint loss is then used to select small-loss samples for model updates. Its effectiveness relies on the assumption that clean samples tend to yield smaller training losses. We hypothesize that this principle may also reduce noise in model representations, rather than only help learning from noisy labels. Under a ranking framework, a consistency loss may therefore denoise the learned representations. In DNA-encoded library screening, the DEL-Ranking preprint combined ranking correction with denoising and consistency objectives[34]. For drug-target affinity regression, BiT-Fusion combined a RankNet-style ranking loss with R-Drop consistency regularization and reported improved performance[27]. However, this approach has not been validated for activity-cliff prediction in small molecules or AMPs.

We therefore propose DualCliff. We use the AC split from AMPCliff[24] to repartition the AC datasets and evaluate AC prediction for both AMPs and small molecules. The AMP evaluation covers datasets for *E. coli*, *S. aureus*, and *P. aeruginosa* [23]. For small molecules, we use three MMP activity-cliff datasets curated by ACGCN: thrombin, the mu-opioid receptor, and melanocortin receptor 4[35]. ACANet also used these three datasets[36]. DualCliff combines an MSE loss, a listwise loss, and a Pairwise Preference Consistency (PPC) loss that we developed from existing methods. PPC is a rank-consistency loss. Conventional consistency losses constrain numerical predictions, whereas PPC constrains the predicted order. We did not use a conventional consistency loss because, without a ranking constraint, it tended to pull predictions toward the mean to reduce the loss. On datasets with a wider label range, this behavior instead reduced Spearman correlation. Across the AMP tasks, DualCliff with ESM2-t12 achieved the highest cross-dataset mean Spearman correlation and Recall@50 among the compared methods. For small molecules, DualCliff with PNA and PPC activated after 120 epochs achieved the highest mean Spearman correlation, while its mean Recall@50 matched ACANet[36]. Compared with HyperSeek[37] and traditional machine-learning methods[24, 35], the main advantage of DualCliff was concentrated in cross-dataset average ranking performance. The leading method still varied across individual datasets and metrics. PPC behavior analysis further showed that initialization asymmetry and delayed activation can improve selected settings, but their effects depend on the dataset and evaluation metric. The current experiments do not establish student-to-teacher denoising as a causal mechanism. Our main contributions are as follows:

1. Under a unified AC split, DualCliff achieved the highest cross-dataset mean Spearman correlation and Recall@50 on three AMP datasets. It also achieved the highest mean Spearman correlation on three small-molecule datasets, while its mean Recall@50 matched ACANet-PNA.
2. We developed PPC to align the relative order of two predictors in the preference-probability space and jointly optimized it with pointwise MSE and a thresholded listwise loss.
3. We characterized the conditions under which PPC is useful. Initialization asymmetry improved the MolCLR-GIN averages but showed target-specific effects. For PNA without pretrained weights, delayed PPC improved selected metrics, but no initialization and activation schedule was best for both average metrics.

## Related Work

ACANet[36] introduced activity-cliff awareness into molecular representation learning. Its ACA loss combines a standard regression loss with a triplet soft-margin loss. Within each mini-batch, the method dynamically constructs an anchor, an activity-similar positive, and an activity-different negative from the observed activity differences. It optimizes only informative triplets that violate the expected ordering of distances in the latent space. The adaptive margin is determined by the true activity differences within each triplet. This design moves samples with similar activity closer in the latent space and places samples with larger activity differences farther from the anchor. We tested a similar metric-learning contrastive constraint in DualCliff. It did not consistently improve ranking performance on the AMP datasets and was therefore excluded from the final model.

HyperSeek[37] uses a protein-guided three-tower architecture. It maps representations of small-molecule ligands, protein pockets, and target protein sequences into Lorentz-model hyperbolic space. Its training objective combines pocket-ligand and sequence-ligand contrastive learning, within-assay listwise affinity ranking, and a hyperbolic hierarchy regularizer on the pocket-ligand representations. The method suggests that hyperbolic geometry may help distinguish structurally similar ligands with different affinities. The AMP datasets used in this study do not include paired target-protein or pocket information. We could therefore not reproduce the three-tower architecture or its protein-guided training objectives. Instead, we adopted its Lorentz-space representation and removed all branches that require protein information. We retained only the ligand-only backbone to construct a hyperbolic ranking variant for AMP sequences. Implementation details are provided in the experimental settings subsection.

JoCoR[33] addresses classification with noisy labels by jointly training two networks on the same mini-batch. Its objective combines the supervised classification loss of each network with a co-regularization term based on symmetric KL divergence. It selects samples according to the joint loss and updates both networks using the small-loss subset. Unlike the disagreement sampling used in Decoupling and Co-teaching+, JoCoR explicitly regularizes the networks toward consistent predictions. DualCliff does not use the small-loss selection strategy of JoCoR. It adopts only the concepts of joint dual-branch optimization and consistency regularization.

Margin-MSE[38] was originally proposed for neural ranking distillation across different architectures. Instead of matching the absolute relevance scores of the teacher and student, it trains the student to match the teacher's score difference for a query-positive-negative triplet. This avoids the requirement that different architectures share the same absolute score baseline. However, the student must still match the numerical scale of the teacher margin. Both PPC and Margin-MSE model relationships across samples, but they use different forms of supervision. The original Margin-MSE uses a pretrained, frozen teacher to provide offline margins. In PPC, two branches are trained concurrently. Their predicted score differences within each mini-batch are converted into sigmoid preference probabilities, and a symmetric MSE is applied to the preference probabilities of the two branches. PPC is therefore inspired by margin-based relation matching, RankNet-style preference probabilities, and online collaborative learning. It is not a direct implementation of Margin-MSE.

# Materials and Methods

Given a mini-batch $B = \{(x_i, y_i)\}$ of size $N$, DualCliff learns two parallel predictors $f_A$ and $f_B$. For sample $i$, the two branches output $p_i^A$ and $p_i^B$, respectively. We use a unified label orientation in both modalities: larger $y_i$ always means stronger activity. For AMP, $y_i = -\log10(\mathrm{MIC})$; for small-molecule benchmarks, the activity labels are also oriented so that larger values indicate stronger potency. At inference time, the final prediction is the branch mean, $\bar{p}_i = (p_i^A + p_i^B)/2$.

## MSE loss

To preserve pointwise activity calibration, both branches are regressed to the ground-truth label using the averaged branch-wise mean squared error:

$$L_{\text{MSE}} = \frac{1}{2N} \sum_{i} [(p_i^A - y_i)^2 + (p_i^B - y_i)^2]$$

This term anchors both predictors to the absolute activity scale and prevents the ranking objectives from drifting away from quantitative potency prediction.

## Pairwise Preference Consistency loss

Instead of forcing the two branches to match raw scores directly, DualCliff aligns them in the pairwise preference space. For every unordered sample pair $(i, j)$ with $i < j$, we define branch-specific preference probabilities

$$q_{ij}^A = \sigma(p_i^A - p_j^A), q_{ij}^B = \sigma(p_i^B - p_j^B)$$

where $\sigma(\cdot)$ is the sigmoid function. If $M = \frac{N(N-1)}{2}$ denotes the number of unordered pairs in the batch, the pairwise preference consistency loss is:

$$L_{\text{PPC}} = \frac{1}{M} \sum_{i<j} (q_{ij}^A - q_{ij}^B)^2$$

PPC therefore encourages the two branches to agree on relative activity ordering even when their score magnitudes are not exactly identical, making the consensus prediction more stable.

## Listwise loss

To emphasize activity-cliff-sensitive ranking, DualCliff constructs a thresholded competitor set for each anchor sample $i$:

$$S_i = \{j | j \neq i, y_i - y_j \geq \delta\}$$

The threshold $\delta$ removes nearly tied examples and keeps the ranking supervision focused on pairs with sufficiently large activity gaps. In the AMP implementation, the threshold can be expressed as $\delta = \log 10(r)$ when it is defined by an MIC fold-change $r$. For branch A, the anchor-wise listwise objective is

$$\ell_i^A = -\log\left(\frac{\exp(\frac{p_i^A}{\tau_L})}{\exp\left(\frac{p_i^A}{\tau_L}\right) + \sum_{k \in S_i} \exp(\frac{p_k^A}{\tau_L})}\right)$$

where $\tau_L$ is a temperature hyperparameter, and branch B is defined analogously.

Let $A = \{i | S_i \neq \emptyset\}$ be the set of valid anchors in the current batch. The final branch-averaged listwise loss is

$$L_{\text{List}} = \frac{1}{2|A|} \sum_{i \in A} (\ell_i^A + \ell_i^B)$$

This formulation makes each high-activity anchor compete against all sufficiently weaker samples in the same batch, so the model is optimized for global rank quality rather than only local pair corrections.

The overall DualCliff objective is the weighted sum of the three terms:

$$L_{\text{DualCliff}} = L_{\text{MSE}} + \lambda_{\text{PPC}} L_{\text{PPC}} + \lambda_{\text{List}} L_{\text{List}}$$

For AMP tasks, we choose $\lambda_{\text{PPC}} = \lambda_{\text{List}} = 1$, $\tau_L = 0.2$, for small molecule tasks, we choose $\lambda_{\text{PPC}} = \lambda_{\text{List}} = 0.1$, $\tau_L = 0.2$. The hyperparameter tuning experiments are in the following section.

## Experimental setting up

### AMP

**Datasets:** Following AMPCliff [24], we adopted the same AC Split protocol and used the three largest AMP subsets from GRAMPA [23]: E. coli, S. aureus, and P. aeruginosa. Each dataset was partitioned into training, validation, and test sets under the same benchmark definition.

**Models:** We evaluated ESM2-t6 and ESM2-t12. For each encoder, one branch $f_A$ was initialized from a pretrained checkpoint and the other branch $f_B$ was initialized randomly.

**Training Procedure:** All AMP experiments used five aligned random seeds, numbered 0 to 4. DualCliff-AMP was trained for 150 epochs with a batch size of 15, a learning rate of 1e-5, an Adam epsilon of 1e-8, and a weight decay of 0.01. We used a cosine-restart learning-rate scheduler and selected checkpoints according to validation Spearman correlation for the branch-averaged prediction. We set both the PPC and listwise loss weights to 1, the listwise temperature to 0.2, and the activity-gap threshold to log10(2) [24]. HypSeek-AMP used the same epoch, batch-size, and learning-rate settings. ACANet-AMP used its released setting of 800 epochs, a batch size of 128, and a learning rate of 1e-4.

### Small Molecule

**Datasets:** For small molecules, we used the three MMP activity-cliff datasets curated by ACGCN [35]: thrombin, μ-opioid receptor, and melanocortin receptor 4. We re-split these datasets under the AC Split principle. All AC pairs were placed in the test set. Molecules that did not form AC pairs with any other sample were assigned to the training or validation set.

**Models:** We used the PNA architecture adopted by ACANet [36] and and a MolCLR-GIN backbone initialized from pretrained weights [39]. For MolCLR-GIN, one branch used the pretrained checkpoint and the other branch used random initialization. For PNA, the two branches used Kaiming-uniform [40] and orthogonal initialization [41] , respectively.

**Training Procedure:** Standard runs used 300 epochs, a batch size of 16, and five aligned random seeds, numbered 0 to 4. We used AdamW with a weight decay of 0.01. The learning rate was 1e-4 for MolCLR-GIN and 1e-5 for PNA. In dual-branch runs, checkpoints were selected according to validation Spearman correlation for the branch-averaged prediction. $\lambda_{\text{PPC}} = \lambda_{\text{List}} = 0.1$. $\tau_L = 0.2$, an activity-gap threshold of 1[36].

### Baselines

**ACANet:** We called the original ACANet training pipeline directly. We only replaced the encoder when needed so that the encoder family matched the target modality.

HyperSeek-AC was implemented as a modality-adapted baseline rather than a full reproduction of the original HyperSeek model. It kept the Euclidean encoder, the nonlinear projection, the Lorentz exponential map, the hyperbolic regression head, and the HypSeek-style listwise ranking loss. It removed the protein sequence, pocket, proxy, and cross-modal contrastive modules. For small molecules, the Euclidean encoder was PNA or MolCLR-GIN; for AMP, the same hyperbolic head was paired with ESM2.

**Classical Models:** For small molecules, the classical baselines used ECFP and MACCS fingerprints, paired with RF, XGBoost, LightGBM, and MLP regressors. For AMP, we used the same model zoo and added handcrafted descriptors (HC) [24] to the ECFP and MACCS feature sets. This design kept the classical comparison consistent across modalities while matching the benchmark settings used in the reference pipelines.

## Main results

## AMP

Table 1. Results of three AMP tasks. The best values are denoted in bold, second best in underline.

| | | *E. coli* | | *S. aureus* | |
|---|---|---|---|---|---|
| regressor | encoder | Spearman | Recall@50 | Spearman | Recall@50 |
| lightgbm | HC | 0.5791±0.0068 | 18.8±2.1679 | 0.5836±0.0087 | 18.0±0.7071 |
| mlp | HC | 0.4626±0.0237 | 10.4±1.8166 | 0.4322±0.0183 | 16.6±0.8944 |
| rf | HC | 0.5781±0.0021 | 17.2±0.8367 | 0.5343±0.0067 | 20.6±0.5477 |
| xgboost | HC | **0.5835±0.0049** | 17.0±2.6458 | 0.5794±0.0068 | 17.0±1.0000 |
| lightgbm | ECFP | 0.3950±0.0052 | 12.2±2.9496 | 0.3735±0.0117 | 12.6±0.5477 |
| mlp | ECFP | 0.3819±0.0321 | 10.2±1.4832 | 0.3481±0.0347 | 12.4±2.1909 |
| rf | ECFP | 0.3968±0.0044 | 9.4±0.5477 | 0.3943±0.0056 | 14.8±0.4472 |
| xgboost | ECFP | 0.4032±0.0055 | 11.2±1.3038 | 0.3927±0.0088 | 12.6±0.5477 |
| lightgbm | MACCS | 0.3462±0.0030 | 5.0±0.0000 | 0.3426±0.0094 | 16.0±1.0000 |
| mlp | MACCS | 0.3414±0.0126 | 7.0±0.7071 | 0.3913±0.0281 | 13.6±0.8944 |
| rf | MACCS | 0.2847±0.0016 | 8.0±0.7071 | 0.3362±0.0072 | 16.0±0.0000 |
| xgboost | MACCS | 0.3362±0.0045 | 7.4±0.8944 | 0.3591±0.0036 | 16.4±0.8944 |
| ACANet | ESM2-t6 | 0.5738±0.0178 | 21.8±1.3038 | **0.6019±0.0106** | 18.6±1.8166 |
| HypSeek-AC | ESM2-t6 | 0.5499±0.0202 | 19.4±0.5477 | 0.5953±0.0152 | 17.6±1.6733 |
| DualCliff | ESM2-t6 | 0.5597±0.0160 | 19.8±3.4928 | 0.5871±0.0273 | 19.2±1.7889 |
| HypSeek-AC | ESM2-t12 | 0.5634±0.0197 | 20.6±1.3416 | 0.5833±0.0130 | 18.2±2.6833 |
| DualCliff | ESM2-t12 | 0.5816±0.0107 | **22.0±1.7321** | 0.5994±0.0187 | **19.6±1.5166** |
| | | *P. aeruginosa* | Average mean metrics | | |
| regressor | encoder | Spearman | Recall@50 | Spearman | Recall@50 |
| lightgbm | HC | 0.4009±0.0072 | 22.2±1.0954 | 0.5212 | 19.7 |
| mlp | HC | 0.1431±0.0328 | 15.4±3.1305 | 0.3460 | 14.1 |
| rf | HC | 0.3746±0.0052 | 21.6±0.5477 | 0.4957 | 19.8 |
| xgboost | HC | 0.4343±0.0189 | 21.8±0.4472 | 0.5324 | 18.6 |
| lightgbm | ECFP | 0.3806±0.0093 | 18.4±0.5477 | 0.3830 | 14.4 |
| mlp | ECFP | 0.3768±0.0403 | **23.8±4.5497** | 0.3689 | 15.5 |
| rf | ECFP | 0.4167±0.0080 | 18.2±0.8367 | 0.4026 | 14.1 |
| xgboost | ECFP | 0.4193±0.0133 | 16.0±0.7071 | 0.4051 | 13.3 |
| lightgbm | MACCS | 0.3580±0.0086 | 16.4±0.5477 | 0.3489 | 12.5 |
| mlp | MACCS | 0.2786±0.0277 | 12.8±0.8367 | 0.3371 | 11.1 |
| rf | MACCS | 0.2893±0.0079 | 15.0±0.7071 | 0.3034 | 13.0 |
| xgboost | MACCS | 0.3635±0.0121 | 17.4±0.5477 | 0.3529 | 13.7 |
| ACANet | ESM2-t6 | 0.4131±0.0224 | 22.2±1.0954 | 0.5296 | 20.9 |
| HypSeek-AC | ESM2-t6 | **0.4722±0.0266** | 23.4±2.7928 | 0.5391 | 20.1 |
| DualCliff | ESM2-t6 | 0.4497±0.0186 | 22.6±0.8944 | 0.5322 | 20.5 |
| HypSeek-AC | ESM2-t12 | 0.4503±0.0146 | 21.6±1.5166 | 0.5323 | 20.1 |
| DualCliff | ESM2-t12 | 0.4370±0.0314 | 22.6±0.5477 | **0.5393** | **21.4** |

Table 1 evaluates whether DualCliff improves ranking across the three AMP tasks. DualCliff with ESM2-t12 provided the strongest cross-dataset balance, with the highest mean Spearman correlation of 0.5393 and mean Recall@50 of 21.4. Its Recall@50 was highest for E. coli and S. aureus, at 22.0 and 19.6, respectively. HypSeek-AC with ESM2-t6 led P. aeruginosa with 23.4. The per-dataset Spearman leaders varied. XGBoost with handcrafted descriptors led E. coli at 0.5835, ACANet led S. aureus at 0.6019, and HypSeek-AC with ESM2-t6 led P. aeruginosa at 0.4722. DualCliff with ESM2-t12 was close to the best Spearman values on E. coli and S. aureus, with gaps of 0.0019 and 0.0025. These results support the cross-dataset balance of DualCliff, while showing that its advantage is not uniform across individual tasks.

## Small molecule

Table 2. The results of three small molecule tasks. The best values are denoted in bold, second best in underline. Note that the two networks of PNA were initialized by kaiming uniform distribution and orthogonal uniform distribution respectively.

| dataset | regressor | encoder | Spearman | Recall@50 |
|---|---|---|---|---|
| **melanocortin_receptor_4** | lightgbm | ecfp | 0.6996 ± 0.0080 | 33.8 ±0.8367 |
| | lightgbm | maccs | 0.6225 ± 0.0067 | 35.4 ±0.5477 |
| | mlp | ecfp | 0.7031 ± 0.0153 | 37.4 ±1.1402 |
| | mlp | maccs | 0.5880 ± 0.0119 | 34.0 ±1.0000 |
| | rf | ecfp | 0.6651 ± 0.0036 | 33.2 ±0.4472 |
| | rf | maccs | 0.6207 ± 0.0059 | 34.0 ±0.0000 |
| | xgboost | ecfp | 0.6909 ± 0.0131 | 35.2 ±0.8367 |
| | xgboost | maccs | 0.6088 ± 0.0114 | 33.8 ±0.4472 |
| | ACANet | pna | 0.7452 ± 0.0170 | 37.6 ±1.1402 |
| | HypSeek-AC | pna | 0.6759 ± 0.0494 | 35.4 ±1.3416 |
| | DualCliff | pna | 0.7628 ± 0.0150 | 37.6 ±1.8166 |
| | **DualCliff(PPC delay 120 epoch)** | **pna** | **0.7650 ± 0.0154** | **38.6 ±1.5166** |
| | ACANet | molclr_gin | 0.5974 ± 0.0342 | 34.0 ±2.1213 |
| | HypSeek-AC | molclr_gin | 0.5298 ± 0.0330 | 30.8 ±1.4832 |
| | **DualCliff** | **molclr_gin** | **0.6198 ± 0.0215** | **34.2 ±1.6432** |
| **mu_opioid_receptor** | lightgbm | ecfp | 0.5495 ± 0.0054 | 25.0 ±0.7071 |
| | lightgbm | maccs | 0.4919 ± 0.0091 | 22.8 ±0.4472 |
| | mlp | ecfp | 0.4986 ± 0.0138 | 23.6 ±1.1402 |
| | mlp | maccs | 0.5040 ± 0.0115 | 22.6 ±0.8944 |
| | rf | ecfp | 0.5326 ± 0.0030 | 22.6 ±0.5477 |
| | rf | maccs | 0.4879 ± 0.0072 | 22.2 ±0.8367 |
| | xgboost | ecfp | 0.5327 ± 0.0064 | 23.6 ±0.5477 |
| | xgboost | maccs | 0.5055 ± 0.0074 | 23.4 ±0.8944 |
| | ACANet | pna | 0.6211 ± 0.0080 | **25.6 ±0.5477** |
| | HypSeek-AC | pna | 0.5794 ± 0.0205 | 24.6 ±0.8944 |
| | DualCliff | pna | 0.6216 ± 0.0084 | 24.6 ±0.8944 |
| | **DualCliff(PPC delay 120 epoch)** | **pna** | **0.6220 ± 0.0052** | 24.8 ±0.4472 |
| | ACANet | molclr_gin | 0.5252 ± 0.0096 | 23.2 ±0.8367 |
| | HypSeek-AC | molclr_gin | 0.5225 ± 0.0152 | 24.0 ±1.2247 |
| | **DualCliff** | **molclr_gin** | **0.5690 ± 0.0087** | **24.0 ±1.2247** |
| **thrombin** | lightgbm | ecfp | 0.6328 ± 0.0051 | 23.6 ±1.1402 |
| | lightgbm | maccs | 0.6086 ± 0.0039 | 18.8 ±1.0954 |
| | mlp | ecfp | 0.6469 ± 0.0083 | 26.8 ±1.0954 |
| | mlp | maccs | 0.5944 ± 0.0119 | 21.0 ±2.4495 |
| | rf | ecfp | 0.5953 ± 0.0024 | 22.6 ±0.5477 |
| | rf | maccs | 0.5960 ± 0.0028 | 20.0 ±0.7071 |
| | xgboost | ecfp | 0.5810 ± 0.0025 | 21.6 ±0.8944 |
| | xgboost | maccs | 0.6082 ± 0.0032 | 20.0 ±0.7071 |
| | ACANet | pna | 0.6508 ± 0.0029 | 28.0 ±0.7071 |
| | HypSeek-AC | pna | 0.6310 ± 0.0188 | 25.8 ±1.7889 |
| | DualCliff | pna | 0.6772 ± 0.0107 | **28.4 ±1.8166** |
| | **DualCliff(PPC delay 120 epoch)** | **pna** | **0.6800 ± 0.0127** | 27.8 ±1.7889 |
| | ACANet | molclr_gin | 0.5678 ± 0.0040 | 22.2 ±1.7889 |
| | HypSeek-AC | molclr_gin | 0.5739 ± 0.0084 | 24.4 ±1.5166 |
| | **DualCliff** | **molclr_gin** | **0.5833 ± 0.0083** | **22.2 ±1.3038** |
| **Average mean metrics** | lightgbm | ecfp | 0.6273 | 27.5 |
| | lightgbm | maccs | 0.5743 | 25.7 |
| | mlp | ecfp | 0.6162 | 29.3 |
| | mlp | maccs | 0.5621 | 25.9 |
| | rf | ecfp | 0.5977 | 26.1 |
| | rf | maccs | 0.5682 | 25.4 |
| | xgboost | ecfp | 0.6015 | 26.8 |
| | xgboost | maccs | 0.5742 | 25.7 |
| | ACANet | pna | 0.6724 | **30.4** |
| | HypSeek-AC | pna | 0.6288 | 28.6 |
| | DualCliff | pna | 0.6872 | 30.2 |
| | **DualCliff(PPC delay 120 epoch)** | **pna** | **0.6890** | **30.4** |
| | ACANet | molclr_gin | 0.5635 | 26.5 |
| | HypSeek-AC | molclr_gin | 0.5421 | 26.4 |
| | **DualCliff** | molclr_gin | **0.5907** | **26.8** |

Table 2 evaluates DualCliff on the three small-molecule tasks. With PNA and PPC activated after 120 epochs, DualCliff achieved the highest Spearman correlation on all three datasets: 0.7650 for melanocortin receptor 4, 0.6220 for the mu-opioid receptor, and 0.6800 for thrombin. The corresponding ACANet-PNA values were 0.7452, 0.6211, and 0.6508. The cross-dataset mean increased from 0.6724 to 0.6890, an absolute gain of 0.0166. Recall@50 showed a different pattern. DualCliff exceeded ACANet on melanocortin receptor 4, at 38.6 versus 37.6, but obtained slightly lower values on the mu-opioid receptor and thrombin, at 24.8 versus 25.6 and 27.8 versus 28.0. Both methods achieved a mean Recall@50 of 30.4. The small-molecule gain is therefore concentrated in rank correlation, while top-ranked recall remains competitive.

# Ablation Study

## AMP

Table 3. Results of three AMP tasks. The best values are denoted in bold, second best in underline. Batching means that the training sequences are first grouped into connected-component clusters by pairwise similarity ≥ 0.9; a greedy "cluster-first" batch sampler then packs whole clusters into each minibatch (splitting a cluster only when it overflows, and backfilling with isolated sequences last), so that highly similar sequences co-occur within a batch and the listwise ranking loss can directly contrast their activity differences (i.e., activity cliffs).

| dataset | encoder | condition | Spearman | Recall@50 |
|---|---|---|---|---|
| ***E. coli*** | ESM2-t6 | MSE single model (AMPCliff) | 0.5369 ± 0.0199 | 17.2000 ± 3.4205 |
| | | MSE Dual model | 0.5501 ± 0.0189 | 19.6000 ± 1.5166 |
| | | **DualCliff w/o listwise loss** | <u>0.5612 ± 0.0202</u> | **21.6000 ± 2.9665** |
| | | DualCliff | 0.5597 ± 0.0160 | 19.8000 ± 3.4928 |
| | | **DualCliff w/ batching** | **0.5768 ± 0.0106** | <u>19.8000 ± 0.4472</u> |
| | ESM2-t12 | MSE single model (AMPCliff) | 0.5777 ± 0.0196 | 18.4000 ± 1.6733 |
| | | MSE Dual model | 0.5781 ± 0.0113 | 20.0000 ± 3.0822 |
| | | **DualCliff w/o listwise loss** | **0.5818 ± 0.0086** | 20.2000 ± 1.9235 |
| | | **DualCliff** | <u>0.5816 ± 0.0107</u> | **22.0000 ± 1.7321** |
| | | DualCliff w/ batching | 0.5683 ± 0.0091 | <u>20.2000 ± 1.4832</u> |
| ***S. aureus*** | ESM2-t6 | MSE single model (AMPCliff) | 0.5891 ± 0.0171 | 17.8000 ± 1.7889 |
| | | MSE Dual model | 0.5799 ± 0.0097 | **19.6000 ± 1.3416** |
| | | **DualCliff w/o listwise loss** | 0.5825 ± 0.0072 | 19.0000 ± 0.7071 |
| | | **DualCliff** | <u>0.5871 ± 0.0273</u> | 19.2000 ± 1.7889 |
| | | **DualCliff w/ batching** | **0.5899 ± 0.0250** | <u>19.6000 ± 1.5166</u> |
| | ESM2-t12 | MSE single model (AMPCliff) | 0.5858 ± 0.0142 | 17.4000 ± 0.5477 |
| | | MSE Dual model | 0.5808 ± 0.0100 | **19.8000 ± 2.2804** |
| | | DualCliff w/o listwise loss | 0.5833 ± 0.0123 | 19.6000 ± 2.3022 |
| | | **DualCliff** | **0.5994 ± 0.0187** | 19.6000 ± 1.5166 |
| | | DualCliff w/ batching | <u>0.5960 ± 0.0119</u> | <u>19.4000 ± 1.1402</u> |
| ***P. aeruginosa*** | ESM2-t6 | MSE single model (AMPCliff) | 0.4241 ± 0.0341 | 21.8000 ± 0.8367 |
| | | MSE Dual model | 0.4206 ± 0.0465 | 22.4000 ± 1.5166 |
| | | DualCliff w/o listwise loss | 0.4213 ± 0.0275 | 22.0000 ± 1.2247 |
| | | **DualCliff** | **0.4497 ± 0.0186** | <u>22.6000 ± 0.8944</u> |
| | | **DualCliff w/ batching** | <u>0.4343 ± 0.0193</u> | **22.8000 ± 0.8367** |
| | ESM2-t12 | MSE single model (AMPCliff) | 0.4223 ± 0.0202 | 22.0000 ± 1.4142 |
| | | MSE Dual model | 0.4130 ± 0.0258 | 22.6000 ± 1.3416 |

| | | DualCliff w/o listwise loss | 0.4251 ± 0.0144 | 23.0000 ± 0.7071 |
|---|---|---|---|---|
| | | **DualCliff** | **0.4370 ± 0.0314** | **22.6000 ± 0.5477** |
| | | DualCliff w/ batching | 0.4289 ± 0.0158 | 22.2000 ± 1.4832 |
| **Average mean metrics** | - | MSE single model (AMPCliff) | 0.5227 | 19.1 |
| | | MSE Dual model | 0.5204 | 20.7 |
| | | DualCliff w/o listwise loss | 0.5259 | 20.9 |
| | | **DualCliff** | **0.5358** | **21.0** |
| | | DualCliff w/ batching | 0.5324 | 20.7 |

Table 3 examines the contributions of dual-branch training, PPC, listwise supervision, and similarity-based batching on the AMP tasks. The single-model MSE baseline achieved mean Spearman correlation of 0.5227 and mean Recall@50 of 19.1. Dual-model MSE increased Recall@50 to 20.7 but reduced Spearman correlation to 0.5204. Adding PPC raised the two averages to 0.5259 and 20.9. The complete DualCliff objective achieved the best joint averages, at 0.5358 and 21.0. Similarity-based batching produced 0.5324 and 20.7. Its effect varied by setting. For example, batching increased E. coli Spearman correlation with ESM2-t6 from 0.5597 to 0.5768, but reduced the ESM2-t12 value from 0.5816 to 0.5683. The average results support the combined loss and ordinary batching, although component effects remain dataset and encoder dependent.

## Small Molecule

Table 4. Ablation results for the three small-molecule tasks. The best values are shown in bold, and the second-best values are underlined. The two PNA branches used Kaiming-uniform and orthogonal initialization, respectively. The batching strategy is the same as that described for Table 3.

| **dataset** | **encoder** | **condition** | **Spearman** | **Recall@50** |
|---|---|---|---|---|
| **melanocortin_receptor_4** | **molclr_gin** | **MSE single model** | 0.5865 ± 0.0266 | 32.2 ± 2.1679 |
| | | **MSE Dual model** | 0.5882 ± 0.0213 | 33.8 ± 1.6432 |
| | | DualCliff w/o listwise loss | 0.5960 ± 0.0080 | 33.4 ± 0.5477 |
| | | **DualCliff** | **0.6198 ± 0.0215** | 34.2 ± 1.6432 |
| | | **DualCliff w/ batching** | 0.5476 ± 0.0641 | **34.8 ± 0.8367** |
| | **PNA** | **MSE Dual model** | **0.7660 ± 0.0126** | 38.2 ± 1.6432 |
| | | DualCliff w/o listwise loss | 0.7641 ± 0.0069 | **39.0 ± 1.2247** |
| | | **DualCliff** | 0.7628 ± 0.0150 | 37.6 ± 1.8166 |
| | | **DualCliff (PPC delay 120 epoch)** | 0.7650 ± 0.0154 | 38.6 ± 1.5166 |
| **mu_opioid_receptor** | **molclr_gin** | MSE single model | 0.5041 ± 0.0055 | 22.4 ± 0.8944 |
| | | MSE Dual model | 0.5637 ± 0.0107 | 23.2 ± 1.3038 |
| | | DualCliff w/o listwise loss | 0.5655 ± 0.0185 | 23.6 ± 1.1402 |
| | | **DualCliff** | **0.5690 ± 0.0087** | **24.0 ± 1.2247** |
| | | **DualCliff w/ batching** | 0.5183 ± 0.0133 | 24.0 ± 0.7071 |
| | **PNA** | MSE Dual model | 0.6170 ± 0.0099 | **25.6 ± 0.8944** |
| | | DualCliff w/o listwise loss | 0.6190 ± 0.0113 | 25.4 ± 1.1402 |
| | | **DualCliff** | 0.6216 ± 0.0084 | 24.6 ± 0.8944 |
| | | **DualCliff (PPC delay 120 epoch)** | **0.6220 ± 0.0052** | 24.8 ± 0.4472 |
| **thrombin** | **molclr_gin** | MSE single model | 0.5796 ± 0.0118 | 24.2 ± 1.0954 |
| | | MSE Dual model | 0.5870 ± 0.0062 | 23.0 ± 1.2247 |
| | | DualCliff w/o listwise loss | 0.5861 ± 0.0089 | 23.8 ± 1.3038 |
| | | **DualCliff** | **0.5833 ± 0.0083** | 22.2 ± 1.3038 |
| | | **DualCliff w/ batching** | 0.5714 ± 0.0066 | **26.2 ± 1.3038** |
| | **PNA** | MSE Dual model | 0.6751 ± 0.0123 | **29.2 ± 1.9235** |
| | | DualCliff w/o listwise loss | 0.6721 ± 0.0069 | 27.2 ± 0.8367 |
| | | **DualCliff** | 0.6772 ± 0.0107 | 28.4 ± 1.8166 |
| | | **DualCliff (PPC delay 120 epoch)** | **0.6800 ± 0.0127** | 27.8 ± 1.7889 |
| **average mean metrics** | **molclr_gin** | MSE single model | 0.5568 | 26.3 |
| | | MSE Dual model | 0.5796 | 26.7 |
| | | DualCliff w/o listwise loss | 0.5825 | 26.9 |
| | | **DualCliff** | **0.5907** | 26.8 |
| | | **DualCliff w/ batching** | 0.5457 | **28.3** |
| | **PNA** | MSE Dual model | 0.6860 | **31.0** |
| | | DualCliff w/o listwise loss | 0.6851 | 30.5 |
| | | DualCliff | 0.6872 | 30.2 |
| | | **DualCliff (PPC delay 120 epoch)** | **0.6890** | 30.4 |

Table 4 compares loss components and molecular backbones. For MolCLR-GIN, mean Spearman correlation increased from 0.5568 with a single MSE model to 0.5796 with two

MSE-trained branches. PPC increased it to 0.5825, and the complete objective reached 0.5907. Similarity-based batching reduced mean Spearman correlation to 0.5457 but increased mean Recall@50 to 28.3. For PNA, the four configurations yielded similar mean Recall@50 values, from 30.2 to 31.0. Their mean Spearman correlations were also close. Dual-model MSE, PPC without listwise loss, complete DualCliff, and delayed PPC achieved 0.6860, 0.6851, 0.6872, and 0.6890, respectively. PPC alone therefore preserved rather than improved the PNA average, while the listwise term and delayed activation produced modest gains. PNA yielded higher metrics than MolCLR-GIN under the reported settings, although the comparison also reflects architectural and initialization differences.

## Hyperparameter tuning

### AMP

Table 5. Sensitivity of ESM2-t6 to the PPC weight $\lambda_{ppc}$, listwise-loss weight $\lambda_{List}$, and listwise temperature $\tau_L$.

| **dataset** | $\lambda_{ppc}$ | $\lambda_{List}$ | $\tau_L$ | Spearman | Recall@50 |
|---|---|---|---|---|---|
| *E. coli* | 0.05 | 0 | 0 | 0.5558 ± 0.0219 | 20.4000 ± 1.9494 |
| | 0.1 | 0 | 0 | 0.5573 ± 0.0185 | 20.0000 ± 1.0000 |
| | 0.5 | 0 | 0 | 0.5556 ± 0.0203 | 20.4000 ± 1.5166 |
| | **1** | **0** | **0** | **0.5612 ± 0.0202** | **21.6000 ± 2.9665** |
| | 1 | 0.1 | 0.2 | 0.5614 ± 0.0189 | **21.0000 ± 1.5811** |
| | 1 | 0.1 | 1 | 0.5612 ± 0.0186 | 21.0000 ± 3.3166 |
| | 1 | 0.5 | 0.2 | 0.5596 ± 0.0172 | 19.0000 ± 2.8284 |
| | 1 | 0.5 | 1 | 0.5631 ± 0.0191 | 20.4000 ± 3.2094 |
| | 1 | 1 | 1 | **0.5642 ± 0.0142** | 19.6000 ± 2.6077 |
| | **1** | **1** | **0.2** | 0.5597 ± 0.0160 | 19.8000 ± 3.4928 |
| *S. aureus* | 0.05 | 0 | 0 | 0.5795 ± 0.0068 | 19.2000 ± 2.1679 |
| | 0.1 | 0 | 0 | **0.5831 ± 0.0092** | **20.2000 ± 2.2804** |
| | 0.5 | 0 | 0 | 0.5788 ± 0.0112 | 19.2000 ± 1.3038 |
| | **1** | **0** | **0** | 0.5825 ± 0.0072 | 19.0000 ± 0.7071 |
| | 1 | 0.1 | 0.2 | **0.5875 ± 0.0113** | 19.0000 ± 1.4142 |
| | 1 | 0.1 | 1 | 0.5825 ± 0.0097 | 19.2000 ± 1.4832 |
| | 1 | 0.5 | 0.2 | 0.5826 ± 0.0255 | **19.6000 ± 0.8944** |
| | 1 | 0.5 | 1 | 0.5793 ± 0.0171 | 18.8000 ± 0.8367 |
| | 1 | 1 | 1 | 0.5855 ± 0.0112 | 18.2000 ± 1.3038 |
| | **1** | **1** | **0.2** | 0.5871 ± 0.0273 | 19.2000 ± 1.7889 |
| *P. aeruginosa* | 0.05 | 0 | 0 | 0.4081 ± 0.0419 | 22.0000 ± 1.5811 |
| | 0.1 | 0 | 0 | 0.4091 ± 0.0372 | 22.2000 ± 1.7889 |
| | 0.5 | 0 | 0 | 0.4123 ± 0.0399 | 22.2000 ± 1.6432 |
| | **1** | **0** | **0** | **0.4213 ± 0.0275** | **22.0000 ± 1.2247** |
| | 1 | 0.1 | 0.2 | 0.4277 ± 0.0280 | 21.6000 ± 0.5477 |
| | 1 | 0.1 | 1 | 0.4205 ± 0.0280 | 22.0000 ± 2.0000 |
| | 1 | 0.5 | 0.2 | <u>0.4444 ± 0.0147</u> | <u>22.2000 ± 0.8367</u> |
| | 1 | 0.5 | 1 | 0.4233 ± 0.0202 | 21.2000 ± 1.0954 |
| | 1 | 1 | 1 | 0.4238 ± 0.0231 | 21.2000 ± 2.0494 |
| | **1** | **1** | **0.2** | **0.4497 ± 0.0186** | **22.6000 ± 0.8944** |
| Average mean metrics | 0.05 | 0 | 0 | 0.4954 | 20.5 |
| | 0.1 | 0 | 0 | 0.4969 | <u>20.8</u> |
| | 0.5 | 0 | 0 | 0.4958 | 20.6 |
| | **1** | **0** | **0** | **0.5017** | **20.9** |
| | 1 | 0.1 | 0.2 | 0.5071 | 20.5 |
| | 1 | 0.1 | 1 | 0.4997 | **20.7** |
| | 1 | 0.5 | 0.2 | 0.5091 | 20.3 |
| | 1 | 0.5 | 1 | 0.5024 | 20.1 |
| | 1 | 1 | 1 | 0.5031 | 19.7 |
| | **1** | **1** | **0.2** | **0.5138** | <u>20.5</u> |

Table 5 evaluates the PPC weight, listwise-loss weight, and listwise temperature for the AMP tasks. Among the PPC-only settings, a PPC weight of 1 achieved the highest mean Spearman correlation of 0.5017 and the highest mean Recall@50 of 20.9. With listwise supervision, a PPC weight of 1, a listwise weight of 1, and a temperature of 0.2 achieved the highest mean Spearman correlation of 0.5138. Its mean Recall@50 was 20.5. This setting also produced the highest P. aeruginosa Spearman correlation of 0.4497. No configuration led both average metrics. The selected coefficients therefore prioritize overall rank correlation while maintaining similar recall.

## Small Molecule

Table 6. Sensitivity of MolCLR-GIN to the PPC weight $\boldsymbol{\lambda_{ppc}}$, listwise-loss weight $\boldsymbol{\lambda_{List}}$, and listwise temperature $\boldsymbol{\tau_L}$.

| dataset | $\lambda_{ppc}$ | $\lambda_{List}$ | $\tau_L$ | Spearman | Recall@50 |
|---|---|---|---|---|---|
| melanocortin_receptor_4 | 0.05 | 0 | 0 | 0.5893 ± 0.0076 | 33.2 ± 1.3038 |
| | 0.5 | 0 | 0 | <u>0.6004 ± 0.0099</u> | **34.0 ± 1.4142** |
| | 1 | 0 | 0 | 0.5938 ± 0.0080 | <u>33.8 ± 1.7889</u> |
| | **0.1** | **0** | **0** | **0.6026 ± 0.0147** | 33.6 ± 1.8166 |
| | 0.1 | 0.5 | 0.2 | 0.6027 ± 0.0250 | 32.4 ± 0.5477 |
| | 0.1 | 0.1 | 1 | 0.5961 ± 0.0123 | 33.6 ± 1.1402 |
| | 0.1 | 0.5 | 1 | 0.6025 ± 0.0116 | <u>33.8 ± 1.9235</u> |
| | 0.1 | 1 | 0.2 | 0.6049 ± 0.0162 | 32.6 ± 0.8944 |
| | 0.1 | 1 | 1 | 0.5971 ± 0.0138 | 33.6 ± 0.8944 |
| | 0.1 | **0.1** | **0.2** | **0.6198 ± 0.0215** | **34.2 ± 1.6432** |
| mu_opioid_receptor | 0.05 | 0 | 0 | 0.5601 ± 0.0101 | **24.6 ± 1.1402** |
| | 0.5 | 0 | 0 | **0.5701 ± 0.0118** | <u>23.8 ± 0.8367</u> |
| | 1 | 0 | 0 | 0.5516 ± 0.0130 | 23.4 ± 1.1402 |
| | **0.1** | **0** | **0** | 0.5656 ± 0.0122 | 23.6 ± 1.1402 |
| | 0.1 | 0.5 | 0.2 | 0.5807 ± 0.0219 | 23.2 ± 1.0954 |
| | 0.1 | 0.1 | 1 | 0.5531 ± 0.0114 | 23.4 ± 1.5166 |
| | 0.1 | 0.5 | 1 | 0.5721 ± 0.0252 | 23.6 ± 1.5166 |
| | 0.1 | 1 | 0.2 | 0.5697 ± 0.0073 | **24.8 ± 0.8367** |
| | 0.1 | 1 | 1 | 0.5666 ± 0.0162 | 23.6 ± 0.8944 |
| | 0.1 | **0.1** | **0.2** | **0.5690 ± 0.0087** | 24.0 ± 1.2247 |
| thrombin | 0.05 | 0 | 0 | 0.5825 ± 0.0068 | 24.0 ± 1.0000 |
| | 0.5 | 0 | 0 | 0.5779 ± 0.0074 | **25.0 ± 1.2247** |
| | 1 | 0 | 0 | 0.5864 ± 0.0016 | 23.6 ± 1.5166 |
| | 0.1 | 0 | 0 | **0.5887 ± 0.0063** | 23.4 ± 1.5166 |
| | 0.1 | 0.5 | 0.2 | 0.5841 ± 0.0089 | 23.2 ± 0.8367 |
| | 0.1 | 0.1 | 1 | 0.5792 ± 0.0130 | 23.4 ± 1.3416 |
| | 0.1 | 0.5 | 1 | **0.5916 ± 0.0117** | **23.6 ± 1.1402** |
| | 0.1 | 1 | 0.2 | 0.5854 ± 0.0074 | 23.4 ± 2.3022 |
| | 0.1 | 1 | 1 | 0.5800 ± 0.0077 | 22.4 ± 1.1402 |
| | 0.1 | **0.1** | **0.2** | 0.5833 ± 0.0083 | 22.2 ± 1.3038 |
| Average mean metrics | 0.05 | 0 | 0 | 0.5773 | 27.3 |
| | 0.5 | 0 | 0 | 0.5828 | **27.6** |
| | 1 | 0 | 0 | 0.5773 | 26.9 |
| | **0.1** | **0** | **0** | **0.5857** | 26.9 |
| | 0.1 | 0.5 | 0.2 | 0.5891 | 26.3 |
| | 0.1 | 0.1 | 1 | 0.5761 | 26.8 |
| | 0.1 | 0.5 | 1 | 0.5887 | **27.0** |
| | 0.1 | 1 | 0.2 | 0.5867 | 26.9 |
| | 0.1 | 1 | 1 | 0.5813 | 26.5 |
| | **0.1** | **0.1** | **0.2** | **0.5907** | 26.8 |

Table 6 evaluates the corresponding coefficients for MolCLR-GIN. The selected setting used a PPC weight of 0.1, a listwise weight of 0.1, and a temperature of 0.2. It achieved the highest reported mean Spearman correlation of 0.5907 and a mean Recall@50 of 26.8. A PPC weight of 0.5 without listwise supervision achieved the highest mean Recall@50 of 27.6, but its mean Spearman correlation was lower at 0.5828. The selected setting therefore favors global rank correlation, while the PPC-only setting favors top-ranked recall.

## PPC Behavior Analysis

Table 7. Effect of the second-branch $f_B$ initialization in DualCliff with MolCLR-GIN. The first branch $f_A$ was initialized from pretrained weights, while the second branch $f_B$ used either random initialization or the same pretrained weights.

| **dataset** | **$f_B$ initialization** | **Spearman** | **Recall@50** |
|---|---|---|---|

| **melanocortin_receptor_4** | **random** | **0.6198 ± 0.0215** | **34.2 ± 1.6432** |
|---|---|---|---|
| | pretrain | 0.6118 ± 0.0141 | 33.2 ± 0.8367 |
| **mu_opioid_receptor** | **random** | **0.5690 ± 0.0087** | **24.0 ± 1.2247** |
| | pretrain | 0.5467 ± 0.0239 | 23.0 ± 0.7071 |
| **thrombin** | random | 0.5833 ± 0.0083 | 22.2 ± 1.3038 |
| | pretrain | **0.5906 ± 0.0093** | **23.4 ± 1.9494** |
| **Average mean metrics** | **random** | **0.5907** | **26.8** |
| | pretrain | 0.5830 | 26.5 |

Table 7 evaluates whether second-branch initialization affects PPC with MolCLR-GIN. A randomly initialized second branch increased mean Spearman correlation from 0.5830 to 0.5907 and mean Recall@50 from 26.5 to 26.8. The improvement occurred on melanocortin receptor 4 and the mu-opioid receptor. On thrombin, two pretrained branches performed better, with Spearman correlation of 0.5906 versus 0.5833 and Recall@50 of 23.4 versus 22.2. Initialization asymmetry therefore improved the cross-dataset averages, but the direction of the effect depended on the target. Because parameter distance was not measured directly, these results support an association with initialization diversity rather than a causal explanation based on parameter divergence.

Table 8. Effects of second-branch $\boldsymbol{f_B}$ initialization and delayed PPC activation in DualCliff with PNA. The first branch $\boldsymbol{f_A}$ always used Kaiming-uniform initialization. The second branch $\boldsymbol{f_B}$ used either Kaiming-uniform or orthogonal initialization.

| **dataset** | **condition** | $f_B$ **initialization** | **Spearman** | **Recall@50** |
|---|---|---|---|---|
| **melanocortin_receptor_4** | **DualCliff** | **kaiming** | **0.7633 ± 0.0107** | **38.0 ± 0.7071** |
| | **DualCliff** | **orthogonal** | **0.7628 ± 0.0150** | **37.6 ± 1.8166** |
| | **DualCliff (PPC delay 80 epoch)** | **kaiming** | 0.7636 ± 0.0077 | 38.0 ± 1.0000 |
| | **DualCliff (PPC delay 80 epoch)** | **orthogonal** | 0.7662 ± 0.0123 | 38.4 ± 1.1402 |
| | **DualCliff (PPC delay 120 epoch)** | **kaiming** | 0.7622 ± 0.0050 | 38.8 ± 1.3038 |
| | **DualCliff (PPC delay 120 epoch)** | **orthogonal** | **0.7650 ± 0.0154** | **38.6 ± 1.5166** |
| | **DualCliff (PPC delay 150 epoch)** | **kaiming** | **0.7637 ± 0.0187** | **38.6 ± 1.5166** |
| | **DualCliff (PPC delay 150 epoch)** | **orthogonal** | 0.7599 ± 0.0087 | 38.0 ± 1.0000 |
| **mu_opioid_receptor** | **DualCliff** | **kaiming** | **0.6245 ± 0.0054** | **25.2 ± 1.3038** |
| | **DualCliff** | **orthogonal** | **0.6216 ± 0.0084** | **24.6 ± 0.8944** |
| | **DualCliff (PPC delay 80 epoch)** | **kaiming** | 0.6097 ± 0.0160 | 24.8 ± 0.8367 |
| | **DualCliff (PPC delay 80 epoch)** | **orthogonal** | 0.6188 ± 0.0051 | 24.4 ± 0.5477 |
| | **DualCliff (PPC delay 120 epoch)** | **kaiming** | 0.6187 ± 0.0083 | 24.8 ± 0.8367 |
| | **DualCliff (PPC delay 120 epoch)** | **orthogonal** | **0.6220 ± 0.0052** | **24.8 ± 0.4472** |
| | **DualCliff (PPC delay 150 epoch)** | **kaiming** | **0.6237 ± 0.0135** | **25.4 ± 1.1402** |
| | **DualCliff (PPC delay 150 epoch)** | **orthogonal** | 0.6175 ± 0.0089 | 24.2 ± 1.3038 |

| **thrombin** | **DualCliff** | **kaiming** | **0.6662 ± 0.0142** | **29.2 ± 0.8367** |
|---|---|---|---|---|
| | **DualCliff** | **orthogonal** | **0.6772 ± 0.0107** | **28.4 ± 1.8166** |
| | **DualCliff (PPC delay 80 epoch)** | **kaiming** | 0.6673 ± 0.0118 | 29.8 ± 1.9235 |
| | **DualCliff (PPC delay 80 epoch)** | **orthogonal** | 0.6735 ± 0.0107 | 28.6 ± 1.8166 |
| | **DualCliff (PPC delay 120 epoch)** | **kaiming** | 0.6737 ± 0.0086 | 29.8 ± 1.6432 |
| | **DualCliff (PPC delay 120 epoch)** | **orthogonal** | **0.6800 ± 0.0127** | **27.8 ± 1.7889** |
| | **DualCliff (PPC delay 150 epoch)** | **kaiming** | **0.6719 ± 0.0156** | **29.0 ± 1.5811** |
| | **DualCliff (PPC delay 150 epoch)** | **orthogonal** | 0.6746 ± 0.0101 | **28.6 ± 1.3416** |
| **Average mean metrics** | **DualCliff** | **kaiming** | 0.6847 | 30.8 |
| | **DualCliff** | **orthogonal** | **0.6872** | 30.2 |
| | **DualCliff (PPC delay 80 epoch)** | **kaiming** | 0.6802 | 30.9 |
| | **DualCliff (PPC delay 80 epoch)** | **orthogonal** | 0.6862 | 30.5 |
| | **DualCliff (PPC delay 120 epoch)** | **kaiming** | 0.6849 | **31.1** |
| | **DualCliff (PPC delay 120 epoch)** | **orthogonal** | **0.6890** | 30.4 |
| | **DualCliff (PPC delay 150 epoch)** | **kaiming** | 0.6864 | 31.0 |
| | **DualCliff (PPC delay 150 epoch)** | **orthogonal** | 0.6840 | 30.3 |

Table 8 evaluates second-branch initialization and PPC activation time for PNA, for which no pretrained checkpoint was used. Without a delay, orthogonal initialization produced a slightly higher mean Spearman correlation than Kaiming-uniform initialization, at 0.6872 versus 0.6847, but a lower mean Recall@50, at 30.2 versus 30.8. Orthogonal initialization with a 120-epoch delay achieved the highest mean Spearman correlation of 0.6890. Kaiming-uniform initialization with the same delay achieved the highest mean Recall@50 of 31.1. The response was not monotonic across datasets or delay times. For example, the orthogonal setting peaked at 120 epochs and declined to 0.6840 at 150 epochs, whereas the Kaiming-uniform setting increased from 0.6847 without a delay to 0.6864 at 150 epochs. Delayed PPC is therefore useful in selected settings, but no single schedule optimizes both metrics.

## Conclusions

This study introduced DualCliff for activity ranking on unseen activity-cliff samples. DualCliff combines absolute-activity regression with ranking learning through pointwise MSE, a thresholded listwise loss, and PPC. PPC aligns the relative order of two predictors in the preference-probability space rather than forcing their numerical outputs to match. Across six datasets, DualCliff achieved the highest cross-dataset mean Spearman correlation and Recall@50 on the AMP tasks. On the small-molecule tasks, its main advantage was a higher mean Spearman correlation, while mean Recall@50 matched ACANet-PNA. The ablation results showed that PPC and listwise supervision can improve ranking in selected settings, but the gains depend on the dataset, backbone, and sampling strategy. The PPC behavior analysis further showed that initialization asymmetry and delayed activation can be beneficial, although no strategy was best across all targets and metrics. PPC should therefore be configured according to the availability of pretrained weights and the training stage. Future work should test this conclusion on more targets and AMP systems, develop adaptive PPC schedules, incorporate protein or membrane context, and quantify predictive uncertainty.